\documentclass[10pt,twocolumn,letterpaper]{article}

\usepackage[pagenumbers]{wacv} 

\definecolor{wacvblue}{rgb}{0.21,0.49,0.74}
\usepackage[pagebackref,breaklinks,colorlinks,allcolors=wacvblue]{hyperref}
\usepackage{algorithm}
\usepackage{algorithmic}

\def\wacvPaperID{827} 
\def\confName{WACV}
\def\confYear{2027}

\title{PuTR-CouT: Counting-by-Tracking in Camera-Trap Image Sequences}

\author{Fagner Cunha, Juan G. Colonna, Eulanda M. dos Santos\\
Federal University of Amazonas\\
Manaus, Amazonas, Brazil\\
{\tt\small \{fagner.cunha, juancolonna, emsantos\}@icomp.ufam.edu.br}
}

\begin{document}
\maketitle
\begin{abstract}
Species identification in camera trap images has been widely studied, but key ecological modeling tasks such as species abundance or density estimation also require counting individual animals. However, the lack of counting labels in most datasets and low frame rates (typically ~1 frame per second) make sequence-level tracking and count estimation particularly challenging. In this work, we present PuTR-CouT, a counting-by-tracking framework built on a transformer-based learned association mechanism for sequence-level animal counting in camera trap images. To address the scarcity of annotated tracking data, we generate synthetic training data by exploiting structural priors, such as static backgrounds and short temporal bursts, to heuristically create pseudo-tracking labels in a weakly supervised manner. The resulting tracker associates detections across frames, using these tracks to estimate per-species counts. We also refine the MaxBoxCount heuristic used by the top solutions of the iWildCam 2021 challenge as a strong baseline, setting the highest score reported to date. When evaluated on the iWildCam 2021 benchmark, our framework PuTR-CouT delivers competitive counting results compared to the improved MaxBoxCount, with the added capability of multi-species predictions and track-level verification.

\end{abstract}
    
\section{Introduction}
\label{sec:intro}

Camera traps are a non-invasive and cost-effective tool for wildlife monitoring
and ecological research \citep{ahumada2013monitoring, he2016visual,
tuia2022perspectives}. However, they produce large volumes of image data
whose manual processing is expensive and time-consuming
\citep{swanson2015snapshot, norouzzadeh2018automatically}. In the last decade,
the computer vision community has investigated diverse approaches to improve
automated information extraction from these data. Much effort has been dedicated
to species identification \citep{norouzzadeh2018automatically,
beery2018recognition, tabak2019machine, willi2019identifying,
schneider2020three}, which enables scientists to model species richness,
occurrence, and distribution \citep{beery2021iwildcam, tuia2022perspectives}.
Yet, certain ecological modeling tasks, such as estimating species abundance
or density, also require counting the number of individuals captured across
image sequences \citep{beery2021iwildcam, moeller2018three,
rowcliffe2013clarifying}.

Counting the number of individuals in a single image can be proxied using an
object detector, such as MegaDetector
\citep{norouzzadeh2021deep,beery2019efficient}. However, most camera trap projects capture multiple
images per event to ensure a clear shot for identifying the photographed animal
\citep{norouzzadeh2018automatically, beery2018recognition}. Therefore,
simply counting bounding boxes can over- or underestimate the actual number of
individuals, as the same individual may appear in multiple images, or different
individuals may appear across different images of the same sequence. This
motivates more sophisticated approaches for sequence-level counting, which may
require tracking or re-identification (re-ID) methods
\citep{beery2021iwildcam}.

Nevertheless, we observe that camera-trap data present several challenges for animal counting.
First, image sequences are typically captured at low frame rates (\eg, 1 frame
per second), causing large displacements between consecutive frames and low
bounding box overlap. This can limit the effectiveness of traditional
multi-object tracking (MOT) methods that rely primarily on geometric cues.
Second, significant appearance changes from varying poses, illumination, or
occlusions complicate re-ID across frames. Third, most camera-trap datasets lack
counting or tracking labels.

The iWildCam 2021 Challenge \citep{beery2021iwildcam} introduced a benchmark
addressing this setting: sequence-level counting at low frame rates without
training count labels, which were only collected for the private test set. Top
challenge solutions relied on simple heuristics, such as taking the maximum
count of bounding boxes (MaxBoxCount) across images, but restricting the
prediction to a single species per sequence. Despite their simplicity, these
heuristic approaches have proven to be strong baselines, with top solutions in
the 2022 challenge continuing to adopt similar strategies \citep{wang2022deep}.
In this work, we revisit MaxBoxCount using stronger detection and
classification models to establish an improved baseline for this counting
challenge.

Recent work further demonstrates the potential of combining temporal
correspondence with individual-level wildlife monitoring.
\citet{schall2026gorillawatch} introduced GorillaWatch, an end-to-end system
integrating detection, multi-object tracking, and individual re-identification
for in-the-wild gorilla monitoring. While their work focuses primarily on video
sequences of a single species and relies on re-ID for counting, our approach is
complementary: we investigate sequence-level counting in low-frame-rate camera
traps, leveraging tracking as an explicit intermediate representation for count
estimation.

In this work, we present a counting-by-tracking framework built on a
transformer-based learned association mechanism for sequence-level animal
counting in camera trap images. To address the scarcity of annotated tracking
data, we generate synthetic training data by exploiting structural priors, such
as static backgrounds and short temporal bursts, to heuristically create
pseudo-tracking labels in a weakly supervised manner. The resulting tracker
associates detections across frames, using these tracks to estimate per-species
counts. When evaluated on the iWildCam 2021 benchmark, our improved version of MaxBoxCount
achieves the highest score reported to date, while our learned
counting-by-tracking framework yields competitive results enabling full
sequence-level tracking and multi-species predictions.

As the main contributions of this paper, we can highlight:
\begin{itemize}
 \item A novel strategy for generating synthetic labeled data for multi-object tracking that takes advantage of camera trap image sequence priors, allowing supervised training of trackers with learned association modules.
 \item The development of a transformer-based counting-by-tracking framework that estimates the number of individuals via tracklets, enabling multi-species predictions across sequences and species classification based on multiple views per individual.
 \item An improved maximum bounding box counting heuristic leveraging state-of-the-art detection and classification models, which achieves the top result reported to date on the iWildCam 2021 benchmark private score.
\end{itemize}

\section{Related Work}
\label{sec:related}

\textbf{Multi-object tracking methods}. Multi-object tracking (MOT) aims to detect all
objects in a video, associate them across frames, and assign a unique identity
to each object in the scene. Modern MOT literature is dominated by the tracking-by-detection paradigm, which separates frame-level object detection from temporal association \citep{zhang2022bytetrack}. Early real-time approaches, such as SORT \citep{wojke2017simple}, rely on linear motion models via Kalman filtering and spatial overlap to perform lightweight data association using the Hungarian algorithm \citep{kuhn1955hungarian}. Subsequent frameworks enhanced tracking robustness by introducing multi-stage matching heuristics to recover low-confidence detections or integrating deep appearance embeddings (ReID) to handle occlusions, as seen in ByteTrack \citep{zhang2022bytetrack} and BoT-SORT \citep{aharon2022bot}. More recently, transformer-based architectures like PuTR \citep{liu2025pure} have replaced rule-based matching with learned attention mechanisms, modeling object associations directly as directed acyclic graphs. Unlike standard vision transformers that tokenize uniform visual patches or full image frames, PuTR treats each detected bounding box as an individual token. Frames without detections generate no tokens (except for an initial sequence token), and spatial-temporal positional encodings inform the causal attention mechanism of both the frame index and spatial location of each box.

\textbf{Counting individuals in camera-trap data}. In recent years, some efforts have been made to address the task of counting animals in camera-trap images. Counting animals in a single image has been framed as a classification problem \citep{norouzzadeh2018automatically} or solved using object detectors to count individuals based on bounding-box counts \citep{norouzzadeh2021deep, Kilima2025}. For counting in sequences of camera-trap images, early solutions relied primarily on simple heuristics, such as selecting the maximum number of bounding boxes across frames, as proposed by top solutions in the iWildCam 2021 and 2022 challenges \citep{beery2021iwildcam, wang2022deep, cunha2026countinganimalscameratrapsimage}. In other domains, animal counting has frequently been formulated as multi-object tracking (MOT), such as pig counting from top-down cameras \citep{chen2020efficient, kim2022embeddedpigcount, huang2023improved}, underwater fishway monitoring \citep{wu2023dynamic}, and sonar video analysis \citep{kay2022caltech}. Recently, \citet{schall2026gorillawatch} introduced the GorillaWatch dataset, combining MOT and individual re-identification (re-ID) to count gorillas in camera-trap videos. However, counting animals in image bursts remains challenging: low frame rates (\eg, 1 frame per second) cause large displacements that degrade traditional MOT methods reliant on motion or geometric cues, while severe appearance changes further complicate tracking.

\section{Improving the MaxBoxCount heuristic as a strong counting
baseline}\label{sec:maxboxcount}

The top solutions in the iWildCam 2021 Challenge adopted a heuristic strategy in
which the number of individuals is estimated as the maximum number of bounding
boxes (MaxBoxCount) in any image across a sequence, assuming that only one
species is present \citep{cunha2026countinganimalscameratrapsimage}. Although this heuristic, by construction, is unable to
predict multiple species within a sequence, this baseline has been shown to be effective on iWildCam 2021. In this section, we revisit MaxBoxCount by updating
its detection and classification modules with improved wildlife monitoring
models introduced since the conclusion of the challenge, providing a strong
baseline for comparison with our proposed counting-by-tracking framework.

For the animal detector, we used
MegaDetectorV1000-redwood\footnote{\url{https://github.com/agentmorris/
MegaDetector/blob/main/docs/release-notes/mdv1000-release.md}} both to generate
bounding boxes for training the classifier and to evaluate the counting
heuristic. The authors recommend a default threshold of $0.3$ or $0.4$ for this
model family. We set the threshold to $0.3$ when generating bounding boxes for
classifier training, as the classifier is more robust to noisy detections, and
to $0.4$ for the counting heuristic as a more conservative choice. For species
classification, we used a Swin-B model with an input resolution of $224 \times
224$ from the Swin Transformer family~\citep{liu2021swin}. Following top challenge solutions, we
adopted an ensemble approach using two models: one trained on full images and
another on cropped bounding boxes.

\textbf{Image preprocessing}. For the crop classifier, we first applied a square
crop centered on each bounding box. Then, for both the full-image and crop
models, the images were resized so that the smallest side was $450$ pixels.
During training, we used a set of standard data augmentation operations: a
random crop with aspect ratio sampled in $[3/4, 4/3]$ and area in $[0.65, 1.0]$,
random horizontal flipping with $50\%$ probability, and
RandAugment~\citep{cubuk2020randaugment} with parameters $N=2$ and $M=9$.
Finally, images were normalized using the ImageNet mean and standard deviation
and resized to $224 \times 224$. During inference, preprocessing consisted only
of normalization and resizing to match the input resolution of the architecture.

\textbf{Implementation details}. The models were initialized with ImageNet-1k pretrained weights from the \texttt{timm} library~\cite{rw2019timm} and trained using the
AdamW~\citep{loshchilov2018decoupled} optimizer with a batch size of 64 for 30 epochs. The initial learning rate was set to $10^{-5}$, and the weight decay was
set to $10^{-7}$. The learning rate was linearly warmed up from 0 to the initial
value for 2 epochs and then decayed to 0 using a cosine
schedule~\citep{he2019bag}. We also employed label
smoothing~\citep{szegedy2016rethinking} with $\epsilon = 0.1$.

\textbf{Classification inference}. Following the top solutions of iWildCam
2021, image-level predictions are computed using a weighted ensemble of
predictions: $0.15 \times \text{full image} + 0.15 \times \text{mirrored full
image} + 0.35 \times \text{bbox} + 0.35 \times \text{mirrored bbox}$.
Sequence-level species predictions are then obtained by averaging the
predictions across all non-empty images in the burst.

\section{Counting-by-Tracking in Camera Trap Image Sequences}

A natural approach to counting animals in a sequence is to track them across
images. In this approach, the detections in each image are associated with the
corresponding detections in other images of the same sequence, framing the
problem as a multi-object tracking task. The number of animals can then be
obtained by counting the number of unique tracks, which we refer to as
counting-by-tracking.

\begin{figure}[t]
  \centering
   \includegraphics[width=\linewidth]{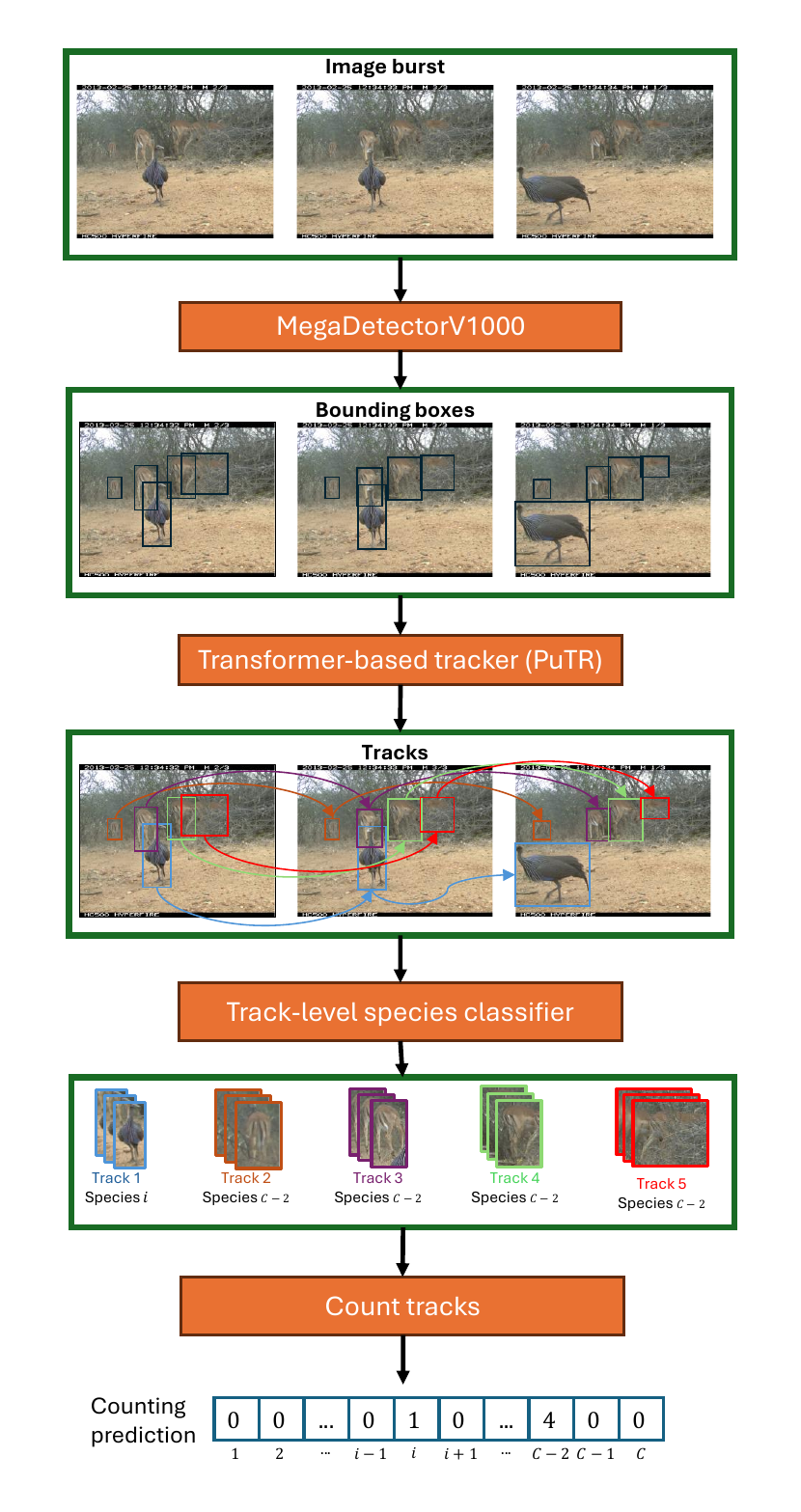}
   \caption{The PuTR-Cout framework first detects bounding boxes using MegaDetectorV1000, connects frame-to-frame detections via a synthetic-trained Transformer tracker to form single-animal tracks, and finally classifies each track independently to output per-species animal counts for the sequence.}
   \label{fig:count-track-pipeline}
\end{figure}

In this section, we explore a counting-by-tracking approach built upon Pure
Transformer (PuTR) \citep{liu2025pure} trained to track animals in camera trap image sequences. Each
track is then classified individually using all associated bounding boxes, and
the number of individuals per species is estimated by counting unique tracks.
Since labeled datasets for multi-object tracking in camera trap image sequences
are usually not available, we propose a method for generating a synthetic
training set with pseudo-labels, enabling weakly supervised training of the
tracking model. \cref{fig:count-track-pipeline} summarizes our approach, which
we call Pure Transformer for Counting-by-Tracking (PuTR-CouT).

\subsection{Generating a synthetic training set for tracking using camera trap
sequence priors}\label{sec:gen_synthetic}

Classical trackers such as SORT and ByteTrack rely on motion models and
geometric overlap. Therefore they do not require a labeled tracking dataset. Trackers that incorporate learned association modules, such as PuTR, require
tracklet annotations for training. Unfortunately, obtaining such labels is
infeasible in most camera trap projects. 

We overcome this limitation, we begin by defining a base case for generating tracking labels as sequences in
which an animal detector produces at most one bounding box per image. Under this
condition, we assume that only a single individual appears throughout the entire
sequence. Although this assumption does not always hold, it provides a
reasonable starting point, and we accept the potential noise introduced by this
simplification.

\begin{figure*}[t]
  \centering
   \includegraphics[width=0.8\linewidth]{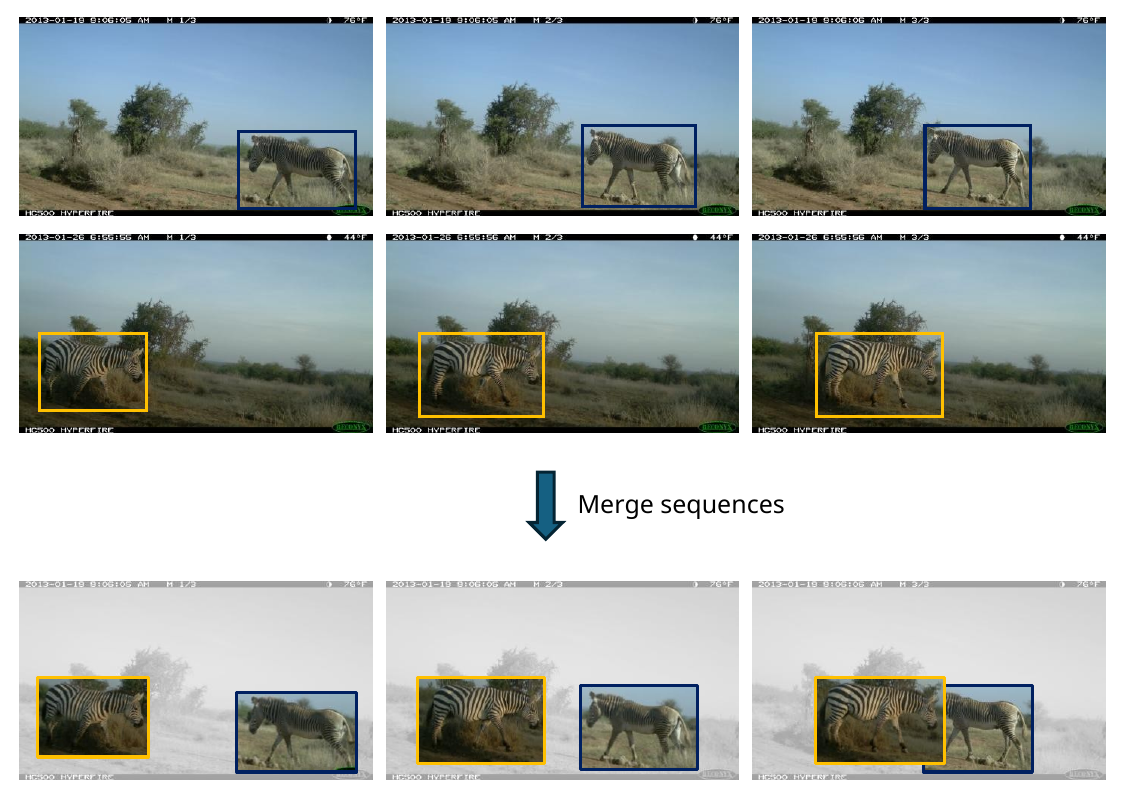}
   \caption{Illustration of the proposed synthetic sequence generation. Base-case tracks captured at the same location and within the same coarse timeslot can be merged to simulate a multi-individual sequence. Each cropped box is treated as a token by PuTR, removing the need to reconstruct composite images.}
   \label{fig:synt_sample}
\end{figure*}

Next, to synthesize sequences containing more than one individual, \ie to simulate animal herds, we leverage the static nature of camera
traps by merging base-case tracks from sequences obtained at the same location,
as shown in \cref{fig:synt_sample}. As the background may gradually
change due to vegetation growth over time, illumination variations throughout
the day, or different weather conditions, to minimize domain shift and visual inconsistencies across merged frames, we restrict the merging
process to sequences captured at the same location and within the same coarse
timeslot\footnote{We define the timeslots as follows: morning from 05:00 to
11:00, afternoon from 11:00 to 17:00, and night for the remaining hours.}:
morning, afternoon, or night. Importantly, under the PuTR formulation, each
input token represents a bounding box rather than the full image. Therefore,
synthetic sequences can be constructed simply by combining cropped bounding
boxes from the original sequences, without explicitly generating synthetic
composite images.

Finally, the number of individuals or different species present in a sequence
may vary across species and locations. Since the number of individuals per
sequence is unknown in our problem setting, we cannot directly replicate
real-world count distributions when generating our synthetic dataset by merging
single-individual sequences. To address this limitation, we employ the
MaxBoxCount heuristic as a proxy for the number of individuals per sequence in a
weakly supervised formulation. By generating synthetic samples that follow this
distribution, we aim to mitigate the impact of distribution shift in the number
of tracks per sequence when evaluating the model on the real test set.

Given the counts produced by MaxBoxCount on the iWildcam 2021 training set, we
used the \texttt{statsmodels} library~\citep{seabold2010statsmodels} to fit a
Negative Binomial distribution (NB), finding the parameters
$\alpha = 1.319389376732665$ and $\mu = 1.0207130544230718$. We then use the
fitted model to sample the number of individuals for each synthetic sequence.
For the number of species, based on the observation that in the Snapshot
Serengeti dataset~\citep{swanson2015snapshot} 95\% of non-empty sequences
contain a single species, we set the probability of observing a single species
to 95\%, two species to 4.9\%, and three species to 0.1\%.
\cref{alg:synthetic_generation} provides a pseudo-code for generating the
synthetic dataset for tracking.

\begin{algorithm}[t]
\caption{Synthetic Sequence Generation}
\label{alg:synthetic_generation}
\begin{algorithmic}[1]

\STATE \textbf{Input:} Training set $\mathcal{T}$, animal detector $Det$,
target number of samples $N$
\STATE \textbf{Output:} Synthetic dataset $\mathcal{D}_{syn}$

\STATE $\mathcal{O} \gets Det(\mathcal{T})$ \COMMENT{Extract detections}
\STATE $\mathcal{C} \gets \text{MaxBoxCount}(\mathcal{O})$
       \COMMENT{Sequence counts}
\STATE Fit Negative Binomial parameters $(\mu,\alpha)$ to $\mathcal{C}$
\STATE $\mathcal{S}_1 \gets
       \{s \in \mathcal{T} \mid \mathcal{C}(s)=1\}$
       \COMMENT{Single-individual sequences}
\STATE $\mathcal{L}_1 \gets \text{listLocations}(\mathcal{S}_1)$
\STATE $\mathcal{D}_{syn} \gets \emptyset$

\FOR{$j=1$ to $N$}
    \STATE Sample $i \sim \text{NB}(\mu,\alpha)$
           \COMMENT{Number of individuals}
    \STATE Sample $k \sim
           \text{Cat}(\{1:0.95,2:0.049,3:0.001\})$
           \COMMENT{Number of species}
    \STATE Sample location $p \sim \mathcal{L}_1$ and timeslot
           $t \sim \{\text{morning, afternoon, night}\}$
    \STATE $\mathcal{Q} \gets \text{Select } i \text{ sequences from }
           \mathcal{S}_1 \text{ matching } (p,t)$
    \STATE \hspace{\algorithmicindent}
           such that total distinct species $\leq k$
    \STATE $d_j \gets \text{mergeSeqs}(\mathcal{Q})$
    \STATE $\mathcal{D}_{syn} \gets
           \mathcal{D}_{syn} \cup \{d_j\}$
\ENDFOR

\STATE \textbf{return} $\mathcal{D}_{syn}$

\end{algorithmic}
\end{algorithm}

\textbf{Implementation details}. To generate bounding boxes, we used
MegaDetectorV1000-redwood with a confidence threshold of $0.4$, following the
improved baseline. We filtered out sequences labeled as empty and generated
75,000 synthetic sequences, of which 35,804 were non-empty. Empty sequences are
ignored during PuTR training and do not contribute to the loss. This number is
comparable to the iWildCam 2021 training set, which contains 36,547 sequences.
Since sequences contain 1--10 images, we aligned images from the first capture
of each sequence and generated synthetic sequences matching the length of the
longest base sequences to preserve temporal consistency.

\subsection{Adapting PuTR for tracking under camera trap
constraints}\label{sec:training_putr}

We build our counting-by-tracking framework on top of the original PuTR
implementation, introducing adaptations to handle the characteristics of
camera-trap image sequences. We retain the original model architecture,
composed of 6 transformer blocks with multi-head self-attention (8 heads),
including temporal and spatial positional encodings and a causal attention
mask. The model is trained to associate each bounding box with one from the
previous frame.

\textbf{Detections}. To maintain consistency with the MaxBoxCount baseline, we
use MegaDetectorV1000-redwood to generate bounding boxes for both training and
inference. This version is based on YOLOv5x6, while the original PuTR uses
YOLOX during inference. We assume that MegaDetectorV1000 produces high-quality
detections for this task, and all detections with confidence above 0.4 are
treated as ground truth during training.

\textbf{Tokenization}. In PuTR, each bounding box is translated into a token.
Although the authors suggest using Re-ID models or DETR-based features to
obtain high-dimensional representations, the embeddings for the tokens are
extracted directly from raw image patches. Given the high appearance variance
across images in camera-trap data, in our framework we choose to adopt
MegaDescriptor~\citep{vcermak2024wildlifedatasets} to generate the embeddings,
a re-ID model designed for wildlife data. Additionally, we evaluate embeddings
extracted from the species classifier trained on iWildCam 2021, the same model
used to identify the species in the framework.

\textbf{Sequence length}. During training, the original PuTR implementation
progressively increases the number of frames used per clip by adjusting the
sampling rate, with the number of frames going from 4 to 128 according to the
epoch. However, while the original implementation considers videos with a
regular frame rate (for example, 30 FPS), this is not the case for camera-trap
image sequences, which usually have 1 FPS and a small number of frames per
sequence. In our dataset, bursts vary from 1 to 10 frames. Therefore, we fix the
sequence length to 10 frames for both training and inference, padding sequences
with fewer frames as needed.

\textbf{Implementation details}. To generate the tokens, the bounding boxes are cropped, resized, and vectorized resulting in 1024-dimensional embeddings, which are precomputed during synthetic dataset generation for computational efficiency. These embeddings are then passed through a linear projection layer to obtain the token representation, matching the PuTR model dimension ($d_{\text{model}} = 512$).

To prevent the model from exploiting ordering biases introduced during synthetic data generation, we randomly permute the bounding boxes within each frame during training. Finally, no image-level data augmentation is applied.

We train PuTR from scratch on the synthetic dataset following the original
training recipe. The model is trained for 7 epochs with a batch size of 4. We
use the AdamW optimizer with an initial learning rate of $0.0002$ and a weight
decay of $0.0005$. Gradients are accumulated over 32 steps, and gradient
clipping is applied with a maximum norm of $1.0$. We also employ a cosine
learning rate decay schedule.

\subsection{Tuning PuTR inference for the counting
task}\label{sec:inference_putr}

PuTR relies on a runtime manager to handle tracklet states and uses the
Hungarian algorithm \citep{kuhn1955hungarian} to associate detections across frames based on the
Transformer affinity matrix. We retain the original runtime manager with minor
adaptations for camera-trap image sequences.

First, we set the detection threshold to $0.4$, so only these detections are
visible to the tracker. Each detection is associated with a detection in the
previous frame; if unmatched, it is initialized as a new track only if its
confidence exceeds $0.5$. These settings maintain consistency with the
thresholds used in the MaxBoxCount baseline and other multi-object trackers in
our experiments.

In the PuTR runtime, a track must accumulate at least three hits to be
confirmed, remaining in a ``new track'' state until then to ensure stability.
While this is suitable for videos with regular frame rates, it is impractical
for camera-trap sequences at 1 FPS, where animals may appear in only a single
frame. Given this constraint, and considering the high quality of MegaDetector
detections, we remove this stabilization requirement.

To improve tracking, PuTR applies a series of geometric and positional
compensations to the affinity matrix produced by the Transformer. However, the
large displacements of animal detections between frames in camera-trap
sequences can make these compensations ineffective or even harmful. Therefore,
we remove them, relying solely on the Transformer predictions for association,
and evaluate this design choice in an ablation study.

We also relax the association thresholds \texttt{ASSO\_THRE1} and
\texttt{ASSO\_THRE2}, which control the association of detections to tracks
based on bounding box IoU and visual affinity, respectively. We set
\texttt{ASSO\_THRE1} to $0.0$, effectively disabling the IoU constraint, and
reduce \texttt{ASSO\_THRE2} from $0.8$ to $0.25$ to account for large appearance
variations across frames. This helps reduce track fragmentation, \ie cases
where a track is incorrectly split into multiple tracks, artificially
increasing the counts.

\begin{table*}
  \centering
  \caption{Main results on iWildCam 2021 for animal counting (MCRMSE).
  We compare the top-3 challenge submissions, the improved MaxBoxCount
  baseline, classical multi-object trackers, and PuTR-CouT. All classifier
  entries are ensembles of the specified architecture. Bold and underlined
  values indicate the best and second-best scores, respectively. Lower is
  better.}
  \label{tab:results_counting_all}
  \begin{tabular}{@{}llllllll@{}}
    \toprule
    Strategy & BBoxes & Classifier & Re-ID & Counting & Classif. &
    Public $\downarrow$ & Private $\downarrow$ \\
    \midrule

    2021 Winning solution \citep{cunha2026countinganimalscameratrapsimage}
      & MDv4 & Eff.-B2 & -- & Boxes & Sequence
      & 0.0300970 & 0.0293278 \\

    2nd place solution
      & MDv3 & Eff.-B2 & -- & Boxes & Sequence
      & 0.0297267 & 0.0293786 \\

    3rd place solution
      & MDv4 & \begin{tabular}[c]{@{}c@{}}
        ResNet152 \\
        + Eff.-B3 \\
        + Eff.-B7
      \end{tabular}
      & -- & Boxes & Sequence
      & 0.0315964 & 0.0308570 \\

    \midrule

    MaxBoxCount (ours)
      & MDv4 & Swin-B & -- & Boxes & Sequence
      & 0.0269689 & 0.0256999 \\

    MaxBoxCount (ours)
      & MDv1000 & Swin-B & -- & Boxes & Sequence
      & 0.0254706 & \textbf{0.0238785} \\

    \midrule

    ByteTrack
      & MDv1000 & Swin-B & -- & Tracks & Sequence
      & 0.0297142 & 0.0282785 \\

    ByteTrack
      & MDv1000 & Swin-B & -- & Tracks & Tracks
      & 0.0285004 & 0.0284988 \\

    \midrule

    BoT-SORT
      & MDv1000 & Swin-B & -- & Tracks & Tracks
      & 0.0279936 & 0.0283493 \\

    BoT-SORT
      & MDv1000 & Swin-B & Swin-B & Tracks & Tracks
      & 0.0272002 & 0.0273195 \\

    BoT-SORT
      & MDv1000 & Swin-B & MegaDescriptor & Tracks & Tracks
      & 0.0270137 & 0.0272918 \\

    \midrule

    PuTR-CouT (ours)
      & MDv1000 & Swin-B & Swin-B & Tracks & Tracks
      & \underline{0.0254377} & \underline{0.0248642} \\

    PuTR-CouT (ours)
      & MDv1000 & Swin-B & MegaDescriptor & Tracks & Tracks
      & \textbf{0.0250704} & 0.0250820 \\

    \bottomrule
  \end{tabular}
\end{table*}

\section{Experiments and Results}

Our experiments were conducted on the iWildCam 2021 dataset using the official
challenge metric, the mean column-wise root mean squared error (MCRMSE). Since
the test set labels are not publicly available, we evaluated our method on the
Kaggle competition platform via late submissions. We report both public and
private scores, each corresponding to approximately 50\% of the test set: the
public score was visible during the competition, whereas the private score,
revealed only after the competition ended, determined the final rankings.

\subsection{Improved MaxBoxCount heuristic results}

To evaluate the extent to which the proposed improvements in
\cref{sec:maxboxcount} impact the performance of the MaxBoxCount
heuristic, we first replace only the classifier component. As shown in
\cref{tab:results_counting_all}, while keeping the bounding box counts from
MegaDetectorV4 as in the winning solution, the error is reduced by 12.4\%, from
$0.0293278$ to $0.0256999$. Notably, the error gap between the first- and
second-place solutions is less than 0.2\%, while the gap to the third place is
approximately 4.9\%. When using MegaDetectorV1000 to generate bounding boxes for
counting, the error decreases further to $0.0238785$, corresponding to an
improvement of approximately 18.5\% relative to the winning solution. These
results demonstrate the effect of the updated models on the heuristic.

\subsection{PuTR-CouT results}

\textbf{Trackers experimental settings}. The improved MaxBoxCount heuristic,
despite its strong performance, is inherently limited to predicting a single
species per sequence (sequence-level classification). In contrast, PuTR-CouT
overcomes this limitation by adopting a count-by-tracking strategy, where each
track is classified independently (track-level classification). Therefore, we
also include ByteTrack and BoT-SORT, two multi-object trackers, in our
experiments to compare PuTR-CouT with approaches that support track-level
classification, enabling a fairer evaluation. Additionally, all trackers use the
same MegaDetectorV1000 bounding boxes as MaxBoxCount, as well as the same
Swin-B-based bounding box classifier for track classification.

For ByteTrack, we set the high track threshold (used as the primary criterion
for selecting bounding boxes for tracking) to $0.4$ to maintain consistency with
MaxBoxCount, while keeping the default low track threshold of $0.1$. For the new
track threshold, we follow the original ByteTrack implementation, setting it to
be $0.1$ higher than the high track threshold (i.e., $0.5$ in this case). For
BoT-SORT, we use the same threshold values. For the additional thresholds in
BoT-SORT, we relax both the proximity and appearance thresholds (to $0.95$ and
$0.5$, respectively) to account for the characteristics of camera trap image
sequences, which often exhibit large displacements and higher appearance
variability across images. Furthermore, since camera traps are static, we
disable the Camera Motion Compensation module.

For BoT-SORT, we conduct experiments both with and without re-ID module to
evaluate their impact. For the re-ID features, we use MegaDescriptor as well as
features produced by the classifier, using the same models employed during the
tokenization process for PuTR.

\textbf{Trackers results}. The results shown in \cref{tab:results_counting_all}
highlight several important trends. When comparing ByteTrack configurations,
track-level classification performs slightly worse than sequence-level
classification on the private score. However, it shows a larger advantage on the
public score. Although ByteTrack yields the weakest performance among all
evaluated trackers, it still surpasses the winning solution.

\begin{figure*}[t]
  \centering
   \includegraphics[width=0.95\linewidth]{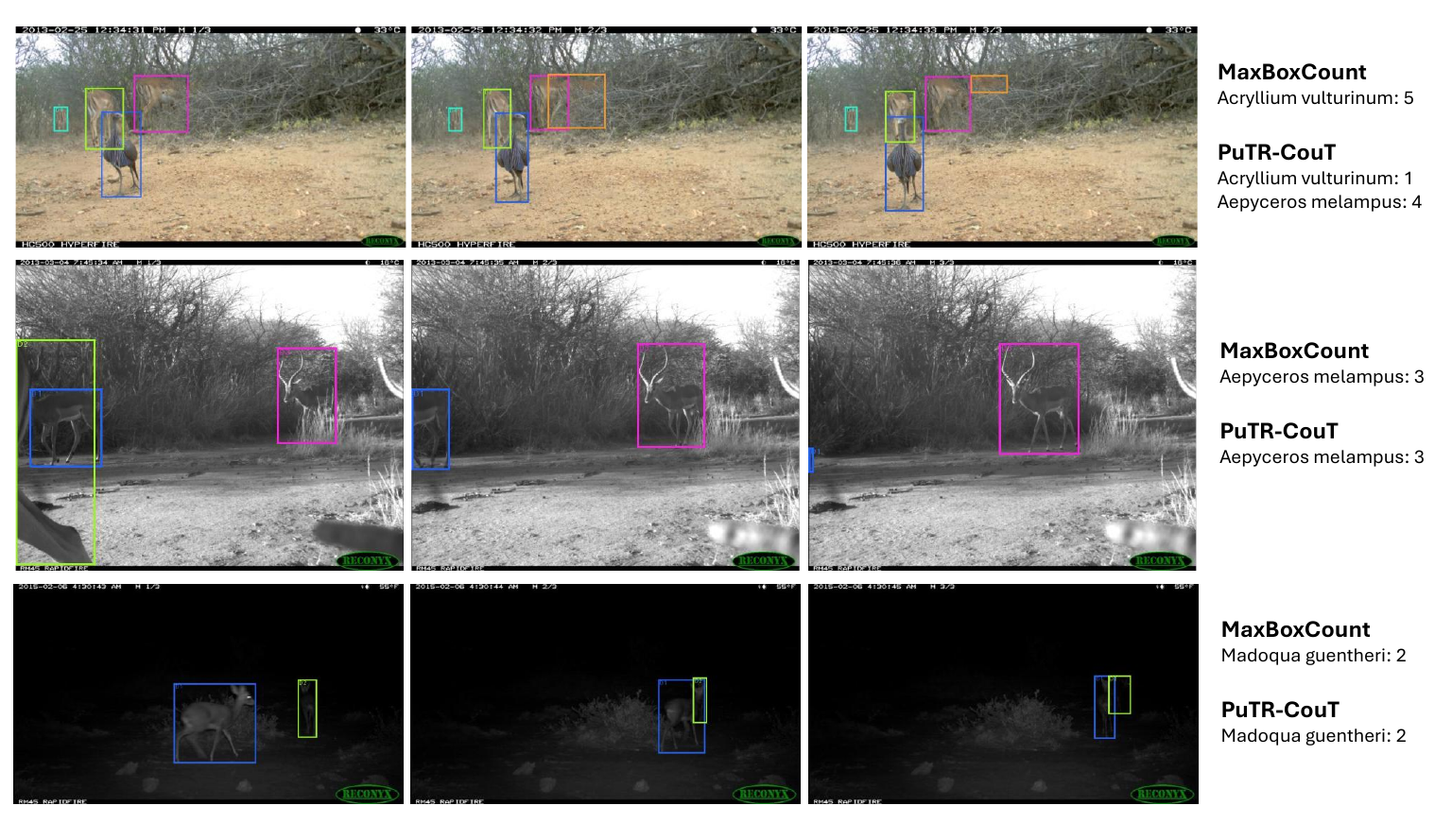}

   \caption{Illustrative examples of counting predictions. Bounding box colors
denote the tracks generated by PuTR-CouT. Only the first three images of each
sequence are displayed. The first row demonstrates a key advantage of PuTR-CouT
over MaxBoxCount in handling multi-species counting scenarios. Best viewed in
color.}
   \label{fig:sample-preds-count}
\end{figure*}

For BoT-SORT, the use of re-ID module leads to improved performance compared to
configuration without re-ID. Finally, PuTR-CouT outperforms all other trackers,
achieving an error of $0.0248642$, corresponding to an improvement of nearly 9\%
over BoT-SORT when using the same re-ID model. In particular, PuTR-CouT with
Swin-B features reduces the error by approximately 15.2\% compared to the
winning solution on the private score. Moreover, PuTR-CouT with MegaDescriptor
achieves the best result on the public score, surpassing even the improved
MaxBoxCount heuristic ($0.0250704$ vs.\ $0.0254706$). Qualitative examples of
the resulting tracks and counting predictions are shown in
\cref{fig:sample-preds-count}.

\subsection{Ablation studies on PuTR-CouT}

\begin{table}[t]
  \centering
  \small
  \setlength{\tabcolsep}{4pt}
  \caption{Effect of reordering camera-trap sequences based on EXIF
  information for the training and test sets, and pre-filtering empty
  bounding boxes. Results are reported on the private test set. Lower is
  better.}
  \label{tab:reordering_results}
  \begin{tabular}{@{}lcccc@{}}
    \toprule
    Tracker &
    \multicolumn{2}{c}{Reorder} &
    Filter &
    Private $\downarrow$ \\
    \cmidrule(lr){2-3}
    & Train & Test & BBoxes & \\
    \midrule

    ByteTrack
      & & & & 0.0282683 \\
    ByteTrack
      & & $\checkmark$ & & 0.0284988 \\

    \midrule

    BoT-SORT (MegaDesc.)
      & & & & 0.0270292 \\
    BoT-SORT (MegaDesc.)
      & & $\checkmark$ & & 0.0272918 \\

    \midrule

    PuTR (MegaDesc.)
      & & & & 0.0256394 \\
    PuTR (MegaDesc.)
      & & $\checkmark$ & & 0.0248314 \\
    PuTR (MegaDesc.)
      & $\checkmark$ & & & 0.0258654 \\
    PuTR (MegaDesc.)
      & $\checkmark$ & $\checkmark$ & & 0.0250820 \\
    PuTR (MegaDesc.)
      & $\checkmark$ & $\checkmark$ & $\checkmark$ & 0.0253143 \\

    \bottomrule
  \end{tabular}
\end{table}

\textbf{Improving tracks}. When constructing the synthetic training set for
tracking, we observed that some image sequences were incorrectly ordered. In
this ablation study, we evaluate the impact of reordering sequences based on
timestamp information extracted from EXIF metadata. For PuTR, we assess all
combinations of applying or not applying reordering during both training and
inference. In contrast, since ByteTrack and BoT-SORT do not involve a training
phase, we evaluate reordering only at inference time.

As shown in~\cref{tab:reordering_results}, reordering slightly degrades performance for both ByteTrack and BoT-SORT, increasing the private error by less than 1\%.
However, for PuTR-CouT, reordering reduces the error by approximately 3\%
compared to the original dataset ordering. This improvement is consistent
regardless of whether reordering is applied during training, highlighting the
effectiveness of sequence reordering for PuTR-CouT. We also investigate
filtering out detections classified as empty before passing them to the tracker.
This strategy, however, leads to an increase in error.

\begin{table}[t]
  \centering
  \small
  \setlength{\tabcolsep}{3.5pt}
  \caption{Ablation of affinity-matrix components in PuTR on camera-trap
  sequences. Results are reported on the private test set. Lower is better.}
  \label{tab:components_results}
  \begin{tabular}{@{}lccccc c@{}}
    \toprule
    Tracker &
    \shortstack[c]{IoU\\comp.} &
    HMIOU &
    HIOU &
    \shortstack[c]{Traj.\\conf.} &
    \shortstack[c]{Det.\\conf.} &
    Private $\downarrow$ \\
    \midrule

    PuTR
      & & & & & & {\bf 0.0250820} \\

    PuTR
      & $\checkmark$ & & & & & 0.0252154 \\

    PuTR
      & & $\checkmark$ & & & & 0.0253234 \\

    PuTR
      & & & $\checkmark$ & & & 0.0253234 \\

    PuTR
      & & & & $\checkmark$ & & 0.0254082 \\

    PuTR
      & & & & & $\checkmark$ & 0.0251428 \\

    \midrule

    PuTR
      & $\checkmark$ & $\checkmark$ & $\checkmark$
      & $\checkmark$ & $\checkmark$ & 0.0253416 \\

    \bottomrule
  \end{tabular}
\end{table}

\textbf{Improving the affinity matrix}. In this ablation study, we evaluate the
impact of disabling the additional components of the affinity matrix introduced
in the original PuTR implementation. As shown in \cref{tab:components_results},
none of these components improve performance in our setting. This finding
supports our hypothesis that incorporating geometric features is challenging for
this task, and suggests that predictions based solely on transformer
representations are already competitive.

\section{Conclusions}

Counting animals in camera trap image sequences is a challenging task, particularly when count labels are unavailable, as is the case for most public datasets. In this work, we introduced PuTR-CouT, a counting-by-tracking approach built on a transformer-based multi-object tracker with learned association. The method was trained using a synthetic dataset generation strategy that exploits structural priors in camera trap sequences, within a weakly supervised setting.

We evaluated our approach against traditional multi-object trackers and the state-of-the-art heuristic MaxBoxCount. The results show that PuTR-CouT achieves
competitive performance while offering key advantages: unlike MaxBoxCount, it inherently supports counting multiple species within the same sequence and
enables direct verification through track analysis.

\section*{Acknowledgments}
This study was financed in part by the Coordenação de Aperfeiçoamento de Pessoal
de Nível Superior - Brasil (CAPES) - Finance Code 001. This work was partially
supported by Amazonas State Research Support Foundation - FAPEAM -  through the
POSGRAD project.

{
    \small
    \bibliographystyle{ieeenat_fullname}
    \bibliography{main}

@String(ICIP = {IEEE Int. Conf. Image Process.})

@String(ICIP  = {ICIP})

@article{norouzzadeh2018automatically,
  title={Automatically identifying, counting, and describing wild animals in
camera-trap images with deep learning},
  author={Mohammad Sadegh Norouzzadeh and Anh Nguyen and Margaret Kosmala
and Alexandra Swanson and Meredith S Palmer and Craig Packer and Jeff Clune},
  journal={Proceedings of the National Academy of Sciences},
  volume={115},
  number={25},
  pages={E5716--E5725},
  year={2018},
  publisher={National Acad Sciences}
}

@article{norouzzadeh2021deep,
  title={A deep active learning system for species identification and counting
in camera trap images},
  author={Norouzzadeh, Mohammad Sadegh and Morris, Dan and Beery, Sara and
Joshi, Neel and Jojic, Nebojsa and Clune, Jeff},
  journal={Methods in Ecology and Evolution},
  volume={12},
  number={1},
  pages={150--161},
  year={2021},
  publisher={Wiley Online Library}
}

@article{Kilima2025,
  author    = {Frank G. Kilima and Shubi Kaijage and Edith Luhanga and Colin
Torney},
  title     = {A YOLOV4 Method for Wild Animal Counting and Behaviour Detection
Using Small-Sized Camera-Trap Image Dataset},
  journal   = {ICTACT Journal on Soft Computing},
  volume    = {15},
  number    = {4},
  pages     = {3722--3728},
  year      = {2025},
  month     = {January},
  publisher = {ICTACT},
  doi       = {10.21917/ijsc.2025.0516}
}

@article{beery2021iwildcam,
  title={The iWildCam 2021 competition dataset},
  author={Beery, Sara and Agarwal, Arushi and Cole, Elijah and Birodkar,
Vighnesh},
  journal={arXiv preprint arXiv:2105.03494},
  year={2021}
}

@inproceedings{wang2022deep,
  title={Deep learning methods for animal counting in camera trap images},
  author={Wang, Yizhen and Zhang, Yang and Feng, Yuan and Shang, Yi},
  booktitle={2022 IEEE 34th International Conference on Tools with Artificial
Intelligence (ICTAI)},
  pages={939--943},
  year={2022},
  organization={IEEE}
}

@inproceedings{chen2020efficient,
  title={Efficient pig counting in crowds with keypoints tracking and
spatial-aware temporal response filtering},
  author={Chen, Guang and Shen, Shiwen and Wen, Longyin and Luo, Si and Bo,
Liefeng},
  booktitle={2020 IEEE International Conference on Robotics and Automation
(ICRA)},
  pages={10052--10058},
  year={2020},
  organization={IEEE}
}

@article{kim2022embeddedpigcount,
  title={EmbeddedPigCount: Pig counting with video object detection and tracking
on an embedded board},
  author={Kim, Jonggwan and Suh, Yooil and Lee, Junhee and Chae, Heechan and
Ahn, Hanse and Chung, Yongwha and Park, Daihee},
  journal={Sensors},
  volume={22},
  number={7},
  pages={2689},
  year={2022},
  publisher={MDPI}
}

@article{huang2023improved,
  title={An improved pig counting algorithm based on YOLOv5 and DeepSORT model},
  author={Huang, Yigui and Xiao, Deqin and Liu, Junbin and Tan, Zhujie and Liu,
Kejian and Chen, Miaobin},
  journal={Sensors},
  volume={23},
  number={14},
  pages={6309},
  year={2023},
  publisher={MDPI}
}

@article{wu2023dynamic,
  title={Dynamic identification and automatic counting of the number of passing
fish species based on the improved DeepSORT algorithm},
  author={Wu, Bilang and Liu, Chunna and Jiang, Furen and Li, Jianyuan and Yang,
Zuobin},
  journal={Frontiers in Environmental Science},
  volume={11},
  pages={1059217},
  year={2023},
  publisher={Frontiers Media SA}
}

@inproceedings{kay2022caltech,
  title={The Caltech Fish Counting dataset: a benchmark for multiple-object
tracking and counting},
  author={Kay, Justin and Kulits, Peter and Stathatos, Suzanne and Deng, Siqi
and Young, Erik and Beery, Sara and Van Horn, Grant and Perona, Pietro},
  booktitle={European Conference on Computer Vision},
  pages={290--311},
  year={2022},
  organization={Springer}
}

@article{ahumada2013monitoring,
  title={Monitoring the status and trends of tropical forest terrestrial
vertebrate communities from camera trap data: a tool for conservation},
  author={Ahumada, Jorge A and Hurtado, Johanna and Lizcano, Diego},
  journal={PloS one},
  volume={8},
  number={9},
  pages={e73707},
  year={2013},
  publisher={Public Library of Science San Francisco, USA}
}

@article{he2016visual,
  title={Visual informatics tools for supporting large-scale collaborative
wildlife monitoring with citizen scientists},
  author={He, Zhihai and Kays, Roland and Zhang, Zhi and Ning, Guanghan and
Huang, Chen and Han, Tony X and Millspaugh, Josh and Forrester, Tavis and
McShea, William},
  journal={IEEE Circuits and Systems Magazine},
  volume={16},
  number={1},
  pages={73--86},
  year={2016},
  publisher={IEEE}
}

@article{tuia2022perspectives,
  title={Perspectives in machine learning for wildlife conservation},
  author={Tuia, Devis and Kellenberger, Benjamin and Beery, Sara and Costelloe,
Blair R and Zuffi, Silvia and Risse, Benjamin and Mathis, Alexander and Mathis,
Mackenzie W and van Langevelde, Frank and Burghardt, Tilo and others},
  journal={Nature communications},
  volume={13},
  number={1},
  pages={1--15},
  year={2022},
  publisher={Nature Publishing Group}
}

@article{moeller2018three,
  title={Three novel methods to estimate abundance of unmarked animals using
remote cameras},
  author={Moeller, Anna K and Lukacs, Paul M and Horne, Jon S},
  journal={Ecosphere},
  volume={9},
  number={8},
  pages={e02331},
  year={2018},
  publisher={Wiley Online Library}
}

@article{rowcliffe2013clarifying,
  title={Clarifying assumptions behind the estimation of animal density from
camera trap rates},
  author={Rowcliffe, J Marcus and Kays, Roland and Carbone, Chris and Jansen,
Patrick A},
  journal={Journal of Wildlife Management},
  year={2013}
}

@inproceedings{beery2018recognition,
  title={Recognition in terra incognita},
  author={Beery, Sara and Van Horn, Grant and Perona, Pietro},
  booktitle={Proceedings of the European Conference on Computer Vision},
  pages={456--473},
  year={2018}
}

@article{swanson2015snapshot,

  title={Snapshot Serengeti, high-frequency annotated camera trap images of 40

mammalian species in an African savanna},

  author={Swanson, Alexandra and Kosmala, Margaret and Lintott, Chris and

Simpson, Robert and Smith, Arfon and Packer, Craig},

  journal={Scientific data},

  volume={2},

  year={2015},

  publisher={Nature Publishing Group}

}

@article{tabak2019machine,
  title={Machine learning to classify animal species in camera trap images:
Applications in ecology},
  author={Tabak, Michael A and Norouzzadeh, Mohammad S and Wolfson, David W and
Sweeney, Steven J and VerCauteren, Kurt C and Snow, Nathan P and Halseth,
Joseph
M and Di Salvo, Paul A and Lewis, Jesse S and White, Michael D and others},
  journal={Meth. in Ecology and Evolution},
  volume={10},
  number={4},
  pages={585--590},
  year={2019},
  publisher={Wiley Online Library}
}

@article{willi2019identifying,
  title={Identifying animal species in camera trap images using deep learning
and citizen science},
  author={Willi, Marco and Pitman, Ross T and Cardoso, Anabelle W and Locke,
Christina and Swanson, Alexandra and Boyer, Amy and Veldthuis, Marten and
Fortson, Lucy},
  journal={Methods in Ecology and Evolution},
  volume={10},
  number={1},
  pages={80--91},
  year={2019},
  publisher={Wiley Online Library}
}

@article{schneider2020three,
  title={Three critical factors affecting automated image species recognition
performance for camera traps},
  author={Schneider, Stefan and Greenberg, Saul and Taylor, Graham W and
Kremer,
Stefan C},
  journal={Ecology and Evolution},
  volume={10},
  number={7},
  pages={3503--3517},
  year={2020},
  publisher={Wiley Online Library}
}

@article{beery2019efficient,
  title={Efficient pipeline for camera trap image review},
  author={Beery, Sara and Morris, Dan and Yang, Siyu},
  journal={arXiv preprint arXiv:1907.06772},
  year={2019}
}

@inproceedings{schall2026gorillawatch,
  title={GorillaWatch: An Automated System for In-the-Wild Gorilla
Re-Identification and Population Monitoring},
  author={Schall, Maximilian and Kn{\"o}fel, Felix Leonard and K{\"o}nig, Noah
Elias and Kubeler, Jan Jonas and von Klinski, Maximilian and Linnemann, Joan
Wilhelm and Liu, Xiaoshi and Schlegelmilch, Iven Jelle and Woyciniuk, Ole and
Schild, Alexandra and others},
  booktitle={2026 IEEE/CVF Winter Conference on Applications of Computer Vision
(WACV)},
  pages={8364--8375},
  year={2026},
  organization={IEEE}
}

@article{seabold2010statsmodels,
  title={Statsmodels: econometric and statistical modeling with python.},
  author={Seabold, Skipper and Perktold, Josef and others},
  journal={scipy},
  volume={7},
  number={1},
  pages={92--96},
  year={2010}
}

@inproceedings{vcermak2024wildlifedatasets,
  title={WildlifeDatasets: An open-source toolkit for animal re-identification},
  author={{\v{C}}erm{\'a}k, Vojt{\v{e}}ch and Picek, Lukas and Adam,
Luk{\'a}{\v{s}} and Papafitsoros, Kostas},
  booktitle={Proceedings of the IEEE/CVF Winter Conference on Applications of
Computer Vision},
  pages={5953--5963},
  year={2024}
}

@inproceedings{cubuk2020randaugment,
  title={Randaugment: Practical automated data augmentation with a reduced
search space},
  author={Cubuk, Ekin D and Zoph, Barret and Shlens, Jonathon and Le, Quoc V},
  booktitle={Proceedings of the IEEE/CVF Conference on Computer Vision and
Pattern Recognition Workshops},
  pages={702--703},
  year={2020}
}

@misc{rw2019timm,
  author = {Ross Wightman},
  title = {PyTorch Image Models},
  year = {2019},
  publisher = {GitHub},
  journal = {GitHub repository},
  howpublished = {\url{https://github.com/rwightman/pytorch-image-models}}
}

@inproceedings{
loshchilov2018decoupled,
title={Decoupled Weight Decay Regularization},
author={Ilya Loshchilov and Frank Hutter},
booktitle={International Conference on Learning Representations},
year={2019},
}

@inproceedings{he2019bag,
  title={Bag of tricks for image classification with convolutional neural
networks},
  author={He, Tong and Zhang, Zhi and Zhang, Hang and Zhang, Zhongyue and Xie,
Junyuan and Li, Mu},
  booktitle={Proceedings of the IEEE/CVF Conference on Computer Vision and
Pattern Recognition},
  pages={558--567},
  year={2019}
}

@inproceedings{szegedy2016rethinking,
  title={Rethinking the inception architecture for computer vision},
  author={Szegedy, Christian and Vanhoucke, Vincent and Ioffe, Sergey and
Shlens, Jon and Wojna, Zbigniew},
  booktitle={Proceedings of the IEEE conference on computer vision and pattern
recognition},
  pages={2818--2826},
  year={2016}
}

@inproceedings{liu2021swin,
  title={Swin transformer: Hierarchical vision transformer using shifted 
windows},
  author={Liu, Ze and Lin, Yutong and Cao, Yue and Hu, Han and Wei, Yixuan and 
Zhang, Zheng and Lin, Stephen and Guo, Baining},
  booktitle={Proceedings of the IEEE/CVF International Conference on Computer 
Vision},
  pages={10012--10022},
  year={2021}
}

@article{kuhn1955hungarian,
  title={The Hungarian method for the assignment problem},
  author={Kuhn, Harold W},
  journal={Naval research logistics quarterly},
  volume={2},
  number={1-2},
  pages={83--97},
  year={1955},
  publisher={Wiley Online Library}
}

@inproceedings{wojke2017simple,
  title={Simple online and realtime tracking with a deep association metric},
  author={Wojke, Nicolai and Bewley, Alex and Paulus, Dietrich},
  booktitle={2017 IEEE international conference on image processing (ICIP)},
  pages={3645--3649},
  year={2017},
  organization={IEEE}
}

@inproceedings{zhang2022bytetrack,
  title={Bytetrack: Multi-object tracking by associating every detection box},
  author={Zhang, Yifu and Sun, Peize and Jiang, Yi and Yu, Dongdong and Weng,
Fucheng and Yuan, Zehuan and Luo, Ping and Liu, Wenyu and Wang, Xinggang},
  booktitle={European conference on computer vision},
  pages={1--21},
  year={2022},
  organization={Springer}
}

@article{aharon2022bot,
  title={BoT-SORT: Robust associations multi-pedestrian tracking},
  author={Aharon, Nir and Orfaig, Roy and Bobrovsky, Ben-Zion},
  journal={arXiv preprint arXiv:2206.14651},
  year={2022}
}

@article{liu2025pure,
  title={Is a Pure Transformer Effective for Separated and Online Multi-Object
Tracking?},
  author={Liu, Chongwei and Li, Haojie and Wang, Zhihui and Xu, Rui},
  journal={ACM Transactions on Multimedia Computing, Communications and
Applications},
  volume={21},
  number={12},
  pages={1--21},
  year={2025},
  publisher={ACM New York, NY}
}

@article{cunha2026countinganimalscameratrapsimage,
      title={Counting Animals in Camera-Traps Image Sequences without Count Labels: Winning Solution to the iWildCam 2021 Challenge}, 
      author={Fagner Cunha and Juan G. Colonna and Eulanda M. dos Santos},
      year={2026},
  journal={arXiv preprint arXiv:2609.03233},
  url={https://arxiv.org/abs/2609.03233},
}
}

\end{document}